\documentclass[10pt,twocolumn,letterpaper]{article}

\usepackage{cvpr}
\usepackage{times}
\usepackage{epsfig}
\usepackage{graphicx}
\usepackage{amsmath}
\usepackage{amssymb}
\usepackage{authblk}

\usepackage[english]{babel}
\usepackage{blindtext}
\usepackage{bbm}
\usepackage{enumitem}
\usepackage{epsfig}
\usepackage{graphicx}
\usepackage{amsmath}
\usepackage{amssymb}
\usepackage[ruled]{algorithm2e}
\usepackage{algorithmic}
\usepackage{multirow}
\usepackage{color}
\usepackage{color, colortbl}
\usepackage{graphicx}
\usepackage{algorithmic}
\usepackage{array}
\usepackage[english]{babel} 
\usepackage{amsmath,amsfonts,amsthm} 
\usepackage{mathtools}

\newcommand{\zs}{\textit{zero-shot}}
\newcommand{\seen}{\textit{seen}}
\newcommand{\unseen}{\textit{unseen}}
\newcommand{\bb}{\mathbf{b}}
\newcommand{\xx}{\mathbf{x}}
\newcommand{\yy}{\mathbf{y}}
\newcommand{\hh}{\mathbf{h}}
\newcommand{\sss}{\mathbf{s}}
\newcommand{\WW}{\mathbf{W}}
\newcommand{\HH}{\mathbf{H}}
\newcommand{\AAA}{\mathbf{A}}
\newcommand{\DD}{\mathbf{D}}

\usepackage[pagebackref=true,breaklinks=true,letterpaper=true,colorlinks,bookmarks=false]{hyperref}

\cvprfinalcopy 

\def\cvprPaperID{****} 

\begin{document}

\title{Zero-Shot Sketch-Image Hashing}
\author[1]{Yuming Shen\thanks{Yuming Shen and Li Liu contributed equally to this work.}
}
\author[2]{Li Liu\footnote[1]
}
\author[3]{Fumin Shen
}
\author[2,1]{Ling Shao
}
\vspace{-3ex}\affil[1]{School of Computing Sciences, University of East Anglia, Norwich, UK}
\affil[2]{Inception Institute of Artificial Intelligence (IIAI), Abu Dhabi, UAE}
\affil[3]{Future Media Center, University of Electronic Science and Technology of China, Chengdu, China}
\affil[ ]{\small \texttt{yuming.shen@uea.ac.uk,  liuli1213@gmail.com,  fumin.shen@gmail.com,  ling.shao@ieee.org}}

\maketitle
\thispagestyle{empty}

\begin{abstract}
	Recent studies show that large-scale sketch-based image retrieval (SBIR) can be efficiently tackled by cross-modal binary representation learning methods, where Hamming distance matching significantly speeds up the process of similarity search. Providing training and test data subjected to a fixed set of pre-defined categories, the cutting-edge SBIR and cross-modal hashing works obtain acceptable retrieval performance. However, most of the existing methods fail when the categories of query sketches have never been seen during training.
	
	In this paper, the above problem is briefed as a novel but realistic zero-shot SBIR hashing task. We elaborate the challenges of this special task and accordingly propose a zero-shot sketch-image hashing (ZSIH) model. An end-to-end three-network architecture is built, two of which are treated as the binary encoders. The third network mitigates the sketch-image heterogeneity and enhances the semantic relations among data by utilizing the Kronecker fusion layer and graph convolution, respectively. As an important part of ZSIH, we formulate a generative hashing scheme in reconstructing semantic knowledge representations for zero-shot retrieval. To the best of our knowledge, ZSIH is the first zero-shot hashing work suitable for SBIR and cross-modal search. Comprehensive experiments are conducted on two extended datasets, i.e., Sketchy and TU-Berlin with a novel zero-shot train-test split. The proposed model remarkably outperforms related works.
\end{abstract}

\vspace{-2ex}\section{Introduction}\label{sec1}
Matching real images with hand-free sketches has recently aroused extensive research interest in computer vision, multimedia and machine learning, forming the term of sketch-based image retrieval (SBIR). Differing the conventional text-image cross-modal retrieval, SBIR delivers a more applicable scenario where the targeted candidate images are conceptually unintelligible but visualizable to user. Several works have been proposed handling the SBIR task by learning real-valued representations~\cite{eitz2010evaluation, skbow, skhog, hu2011bag, siamese, saavedra2014sketch, sketchy, bmvcsong, sketchatt, yu2016sketch, san}. As an extension of conventional data hashing techniques~\cite{gionis1999similarity,seh,itq,dvb}, cross-modal hashing~\cite{cmssh, cmfh, scm, seph, cvh, dcmh, cdq, dvsh} show great potential in retrieving heterogeneous data with high efficiency due to the computationally cheap Hamming space matching, which is recently adopted to large-scale SBIR in~\cite{dsh} with impressive performance. Entering the era of big data, it is always feasible and appreciated to seek binary representation learning methods for fast SBIR.

\begin{figure}[t]\vspace{-3ex}
	\begin{center}
		\includegraphics[width=0.99\linewidth]{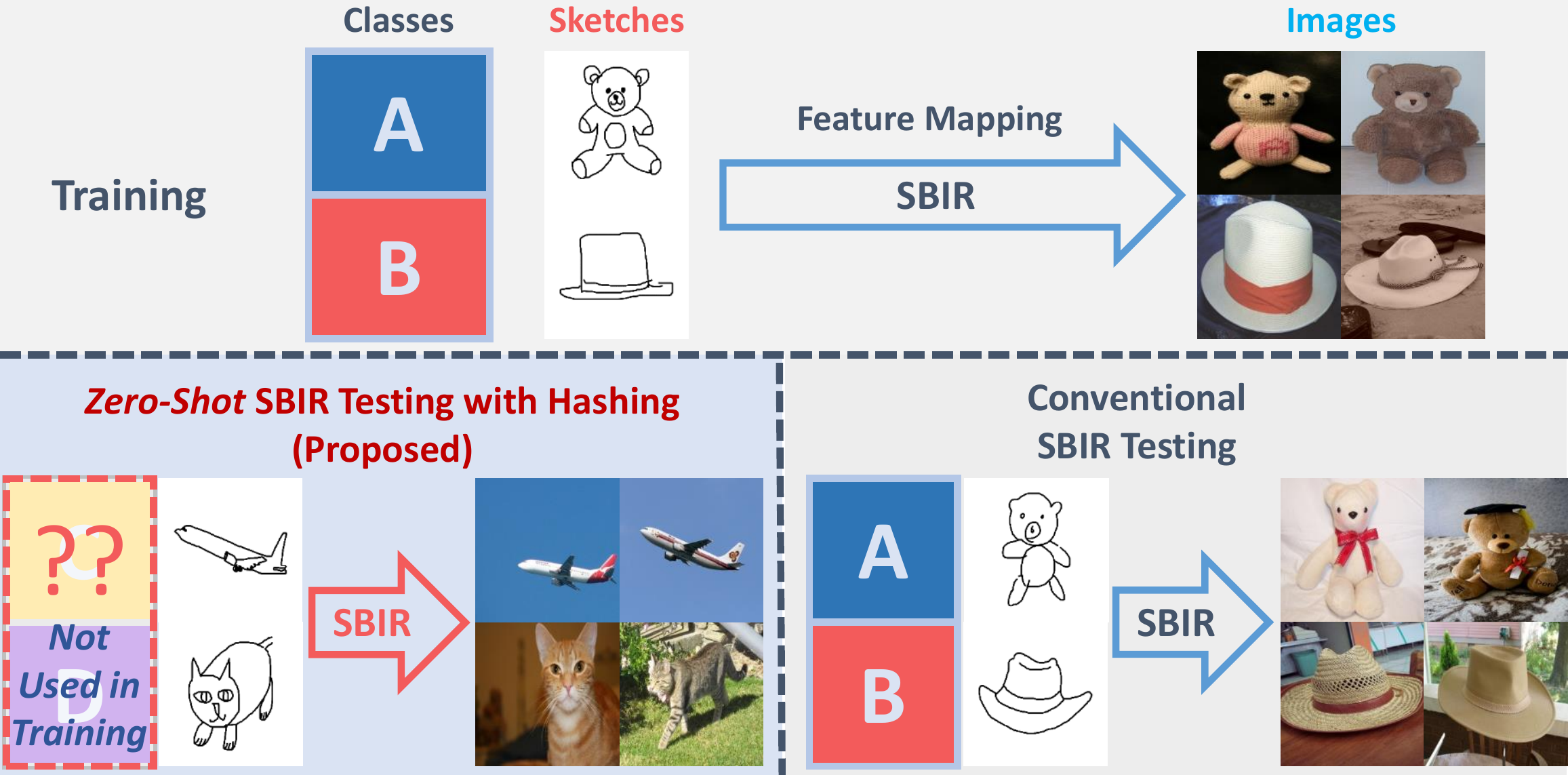}
	\end{center}
	\caption{In conventional SBIR and cross-modal hashing (\textbf{bottom right}), the categories of training data include the ones of test data, marked as \textit{`A'} and \textit{`B'}. For our \zs~task  (\textbf{bottom left}), training data are still subjected to class \textit{`A'} and \textit{`B'}, but test sketches and images are coming from other categories, \ie, \textit{`plane'} and \textit{`cat'} in this case. Note that data labels are not used as test inputs and the test data categories shall be unknown to the learning system.}
	\label{fig1}
	\vspace{-2ex}
\end{figure}

However, the aforementioned works suffer from obvious drawbacks. Given a fixed set of categories of training and test data, these methods successfully manage to achieve sound SBIR performance, which is believed to be a relatively easy task as the visual knowledge from all concepts has been explored during parameter learning, while in a real-life scenario, there is no guarantee that the training data categories cover all concepts of potential retrieval queries and candidates in the database. An extreme case occurs when test data are subjected to an absolutely different set of classes, excluding the trained categories. Unfortunately, experiments show that existing cross-modal hashing and SBIR works generally fail on this occasion as the learned retrieval model has no conceptual knowledge about what to find.

Considering both the \textbf{train-test category exclusion} and \textbf{retrieval efficiency}, a novel but realistic task yields \zs~SBIR hashing. Fig.~\ref{fig1} briefly illustrates the difference between our task and conventional SBIR task. In conventional SBIR and cross-modal hashing, the categories of training data include the ones of test data, marked as \textit{`A'} and \textit{`B'} in Fig.~\ref{fig1}. On the other hand, for the \zs~task, though training data are still subjected to class \textit{`A'} and \textit{`B'}, test sketches and images are coming from other categories, \ie, \textit{`plane'} and \textit{`cat'} in this case. In the rest of this paper, we denote the training and test categories as \seen~and \unseen~classes, since they are respectively known and unknown to the retrieval model.

Our \zs~SBIR hashing setting is a special case of \zs~learning in inferring knowledge out of the training samples. However, existing works basically focus on single-modal \zs~recognition~\cite{cmt,sse, jlse, sae}, and are not suitable for efficient image retrieval. In~\cite{zsh}, an inspiring \zs~hashing scheme is proposed for large-scale data retrieval. Although \cite{zsh} suggests a reasonable \zs~train-test split close to Fig.~\ref{fig1} for retrieval experiments, it is still not capable for cross-modal hashing and SBIR.

Regarding the drawbacks and the challenging task discussed above, a novel \zs~sketch-image hashing (ZSIH) model is proposed in this paper, simultaneously delivering (1) cross-modal hashing, (2) SBIR and (3) \zs~learning. Leveraging state-of-the-art deep learning and generative hashing techniques, 
we formulate our deep network according to the following problems and themes:
\begin{enumerate}[label=(\alph*)]
	\item\label{pa}\vspace{-1.2ex} Not all regions in an image or sketch are informative for cross-modal mapping.
	\item\label{pb}\vspace{-1.2ex} The heterogeneity between image and sketch data needs to be mitigated during training to produce unified binary codes for matching.
	\item\label{pc}\vspace{-1.2ex} Since visual knowledge alone is inadequate for \zs~SBIR hashing, a back-propagatable deep hashing solution transferring semantic knowledge to the \unseen~classes is desirable.
\end{enumerate}\vspace{-1ex}
The contributions of this work are summarized as follows:
\begin{itemize}
	\item\vspace{-1.2ex} To the best of our knowledge, ZSIH is the first \zs~ hashing work for large-scale SBIR.
	\item\vspace{-1.2ex} We propose an end-to-end three-network structure for deep generative hashing, handling the train-test category exclusion and search efficiency with attention model, Kronecker fusion and graph convolution.
	\item\vspace{-1.2ex} The ZSIH model successfully produces reasonable retrieval performance under the \zs~setting, while existing methods generally fail.
\end{itemize}
\textbf{Related Works.} General cross-modal binary representation learning methods~\cite{cmssh,cmfh,scm,cvh,imh,cmnn,seph,dcmh,dvsh,qch,Liong_2017_ICCV,tvdb,yang2017pairwise,cdq,Mandal_2017_CVPR} target to map large-scale heterogeneous data with low computational cost. SBIR, including fine-grained SBIR, learns shared representations to specifically mitigate the expressional gap between hand-crafted sketches and real images~\cite{eitz2010evaluation,skbow,skhog,hu2011bag,bmvcpang,parui2014similarity,siamese,saavedra2014sketch,sketchy,bmvcsong,sketchatt,3dshape,yu2016sketch,san,zhen2012co,zhou2012sketch}, while the efficiency issue is not considered. \textit{Zero-shot}~learning~\cite{devise,sae,sse,jlse,cmt,norouzi2013zero,changpinyo2016synthesized,akata2015evaluation,xian2016latent,al2016recovering,lampert2014attribute,demirel2017attributes,changpinyo2016predicting,Jiang_2017_ICCV,Zhang_2017_CVPR,Ding_2017_CVPR,Li_2017_CVPR,Xu_2017_CVPR,Karessli_2017_CVPR,Ye_2017_CVPR,longyangaaai} is also related to our work, though it does not originally focus on cross-modal retrieval. Among the existing researches, zero-shot hashing (ZSH)~\cite{zsh} and deep sketch hashing (DSH)~\cite{dsh} are the two closest works to this paper. DSH~\cite{dsh} considers fast SBIR with deep hashing technique, but it fails to handle the \zs~setting. ZSH~\cite{zsh} extends the traditional \zs~task to a retrieval scheme.

\section{The Proposed ZSIH Model}
\begin{figure*}
	\begin{center}
		\includegraphics[width=0.97\textwidth, height=0.32\textheight]{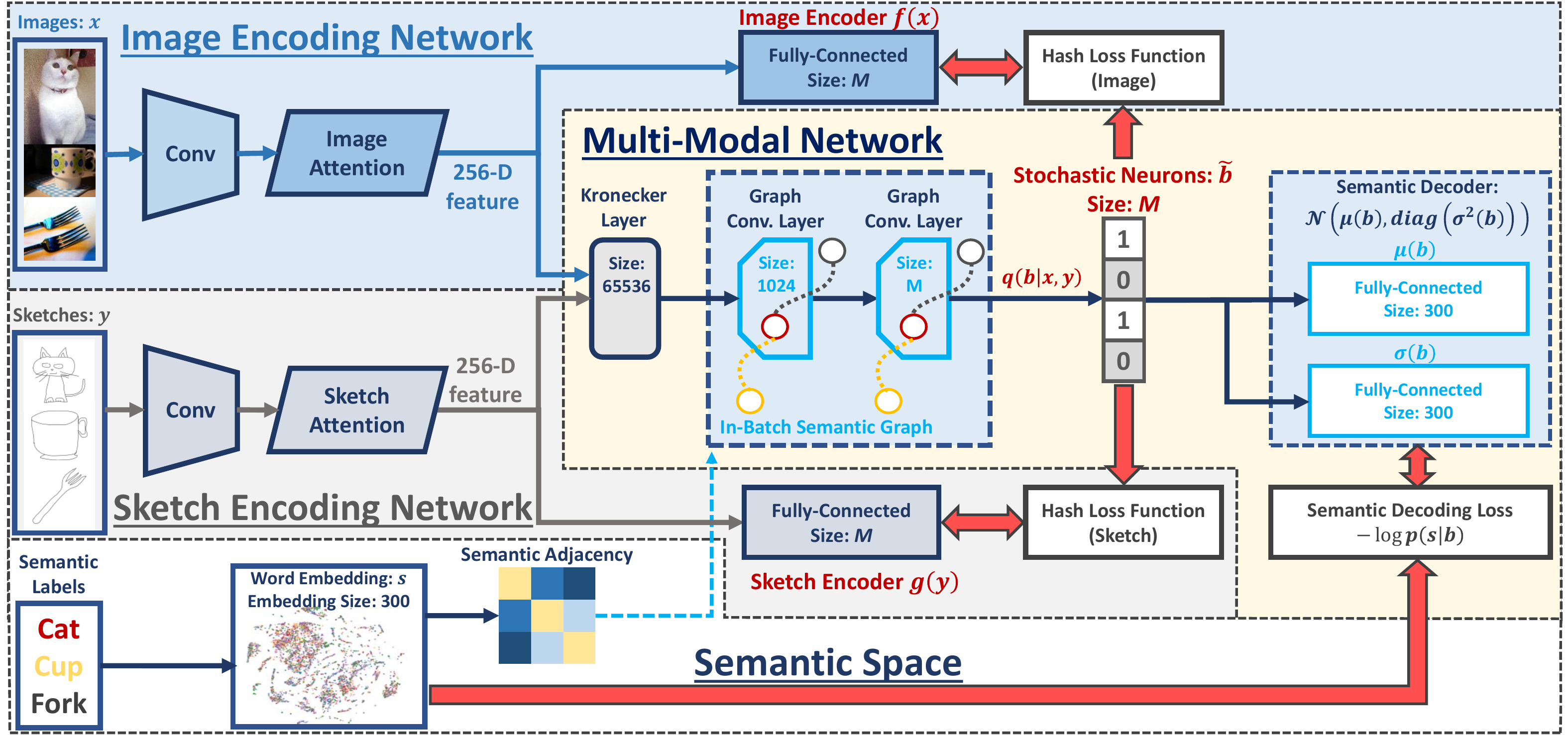}
	\end{center}
	\caption{The deep network structure of ZSIH. The image (in light blue) and sketch encoding network (in grey) act as hash function for the respective modality with attention models~\cite{sketchatt}. The multi-modal network (in canary yellow) only functions during training. Sketch-image representations are fused by a Kronecker layer~\cite{kron}. Graph convolution~\cite{gcn} and generative hashing techniques are leveraged to explore the semantic space for \zs~SIBR hashing. \textbf{Network configurations} are also provided here.}
	\label{fig2}
	\vspace{-2ex}
\end{figure*}
This work focuses on solving the problem of hand-free SBIR using deep binary codes under the \zs~setting, where the image and sketch data belonging to the \seen~categories are only used for training. The proposed deep networks are expected to be capable for encoding and matching the \unseen~sketches with images, categories of which have never appeared during training.

We consider a multi-modal data collection  $\mathcal{O}^c=\{\mathbf{X}^c,\mathbf{Y}^c\}$ from \seen~categories $\mathcal{C}^c$ covering both real images $\mathbf{X}^c=\{\mathbf{x}^c_i\}_{i=1}^N$ and sketch images $\mathbf{Y}^c=\{\mathbf{y}^c_i\}_{i=1}^N$ for training, where $N$ indicates the set size. For the simplicity of presentation, it is assumed that image and sketch data with the same index $i$, \ie, $\mathbf{x}_i^c$ and $\mathbf{y}_i^c$ share the same category label. Additionally, similar to many conventional \zs~learning algorithms, our model requires a set of semantic representations $\mathbf{S}^c=\{\mathbf{s}^c_i\}_{i=1}^N$ in transferring supervised knowledge to the \unseen~data. The aim is to learn two deep hashing functions $f\left(\cdot\right)$ and $g\left(\cdot\right)$ for images and sketches respectively. Given a set of image-sketch data $\mathcal{O}^u=\{\mathbf{X}^u,\mathbf{Y}^u\}$ belonging to the \unseen~categories $\mathcal{C}^u$ for test, the proposed deep hashing functions encode these \unseen~data into binary codes, \ie, $ f:\mathbb{R}^d\rightarrow \{0,1\}^M, g:\mathbb{R}^d\rightarrow \{0,1\}^M$, where $d$ refers to the original data dimensionality and $M$ is the targeted hash code length. 
Concretely, as the proposed model handles SBIR under the \zs~setting, there should be no intersection between the \seen~categories for training and the \unseen~classes for test, \ie, $\mathcal{C}^c\bigcap\mathcal{C}^u=\varnothing$. 
\subsection{Network overview}\label{sec31}
The proposed ZSIH model is an end-to-end deep neural network for \zs~sketch-image hashing. 
The architecture of ZSIH is illustrated in Fig.~\ref{fig2}, which is composed of three concatenated deep neural networks, \ie, the image/sketch encoders and the multi-modal network, 
to tackle the problems discussed above.

\vspace{-1ex}\subsubsection{Image/sketch encoding networks} 
As is shown in Fig.~\ref{fig2}, the networks with light blue and grey background refer to the binary encoders $f\left(\cdot\right)$ and $g\left(\cdot\right)$ for images and sketches respectively. An image or sketch is firstly rendered to a set of corresponding convolutional layers to produce a feature map, and then the attention model mixes informative parts into a single feature vector for further operation. The AlexNet~\cite{alexnet} before the last pooling layer is built to obtain the feature map. We introduce the attention mechanism in solving issue~\ref{pa}, of which the structure is close to \cite{sketchatt} with weighted pooling to produce a 256-D feature. Binary encoding is performed by a fully-connected layer taking input from the attention model with a $\mathtt{sigmoid}$ nonlinearity. During training, $f\left(\cdot\right)$ and $g\left(\cdot\right)$ are regularized by the output of the multi-modal network, so these two encoders are supposed to be able learn modal-free representations for \zs~sketch-image matching.

\vspace{-0.5ex}\subsubsection{Multi-modal network as code learner}\label{sec312}
The multi-modal network only functions during training. It learns the joint representations for sketch-image hashing, handling the problem~\ref{pb} of modal heterogeneity. One possible solution for this is to introduce a fused representation layer taking inputs from both image and sketch modality for further encoding. Inspired by Hu \etal~\cite{kron}, we find the Kronecker product fusion layer suitable for our model, which is discussed in Sec.~\ref{seckron}. Shown in Fig.~\ref{fig2}, the Kronecker layer takes inputs from the image and sketch attention model, and produces a single feature vector for each pair of data points. 
We index the training images and sketches in a coherent category order. Therefore the proposed network is able to learn compact codes for both images and sketches with clear categorical information.

However, simply mitigating the model heterogeneity does not fully solves the challenges in ZSIH. As is mentioned in problem~\ref{pc}, for \zs~tasks, it is essential to leverage the semantic information of training data to generalize knowledge from the \seen~categories to the \unseen~ones. Suggested by many \zs~learning works~\cite{sae,devise,zsh}, the semantic representations, \eg, word vectors~\cite{wv}, implicitly determine the category-level relations between data points from different classes. Based on this, during the joint code learning process, we novelly enhance the hidden neural representations by the semantic relations within a batch of training data using the graph convolutional networks (GCNs)~\cite{gcn2,gcn}. It can be observed in Fig.~\ref{fig2} that two graph convolutional layers are built in the multi-modal network, successively following the Kronecker layer. 
In this way, the in-batch data points with strong latent semantic relations are entitled to interact during gradient computation. Note that the output length of the second graph convolutional layer for each data point is exactly the target hash code length, \ie, $M$. The formulation of the semantic graph convolution layer is given in Sec.~\ref{secgcn}.

To obtain binary codes as the supervision of $f\left(\cdot\right)$ and $g\left(\cdot\right)$, we introduce the stochastic generative model~\cite{sgh} for hashing. A back-propagatable structure of stochastic neurons is built on the top of the second graph convolutional layer, producing hash codes. Shown in Fig.~\ref{fig2}, a decoding model is topped on the stochastic neurons, reconstructing the semantic information. By maximizing the decoding likelihood with gradient-based methods, the whole network is able to learn semantic-aware hash codes, which also accords our perspective of issue~\ref{pc} for \zs~sketch-image hashing. We elaborate on this design in Sec.~\ref{secsgh}~and~\ref{secobj}.


\subsection{Fusing sketch and image with Kronecker layer}\label{seckron}
Sketch-image feature fusion plays an important role in our task as is addressed in problem~\ref{pb} of Sec.~\ref{sec1}. An information-rich fused neural representation is in demand for accurate encoding and decoding. To this end, we utilize the recent advances in Kronecker-product-based feature learning~\cite{kron} as the fusion network. Denoting the attention model outputs of a sketch-image pair $\{\yy,\xx\}$ from the same category as $\hh^{(sk)}\in\mathbb{R}^{256}$ and $\hh^{(im)}\in\mathbb{R}^{256}$, a non-linear data fusion operation can be derived as
\vspace{-1ex}\begin{equation}\vspace{-1ex}
	\mathcal{W}\times \prescript{}{1}{\hh^{(sk)}}\times\prescript{}{3}{\hh^{(im)}}.
\end{equation}
Here $\mathcal{W}$ is a third-order tensor of fusion parameters and $\times$ denotes tensor dot product. We use the left subscript to indicate on which axis tensor dot operates.
De-compositing $\mathcal{W}$ with Tucker decomposition~\cite{tucker1966some}, the fused output of the Kronecker layer $\hh^{(kron)}$ in our model is derived as
\vspace{-0.5ex}\begin{equation}\vspace{-0.5ex}
	\hh^{\left(kron\right)} = \delta\big((\hh^{\left(sk\right)}\WW^{\left(sk\right)})\otimes(\hh^{\left(im\right)}\WW^{\left(im\right)})\big),
\end{equation}
resulting in a 65536-D feature vector. Here $\otimes$ is the Kronecker product operation between two tensors, and $\WW^{\left(sk\right)},\WW^{\left(im\right)}\in\mathbb{R}^{256\times 256}$ are trainable linear transformation parameters. 
$\delta\left(\cdot\right)$ refers to the activation function, which is the $\mathtt{ReLU}$~\cite{relu} nonlinearity for this layer.

Kronecker layer~\cite{kron} is supposed to be a better choice in feature fusion for ZSIH than many conventional methods such as layer concatenation or factorized model~\cite{mfb}. This is because the Kronecker layer largely expands the feature dimensionality of the hidden states with a limited number of parameters, and thus consequently stores more expressive structural relation between sketches and images.
\subsection{Semantic-relation-enhanced hidden representation with graph convolution}\label{secgcn}
In this subsection, we describe how the categorical semantic relations are enhanced in our ZSIH model using GCNs. Considering a batch of training data $\{\xx_i,\yy_i,\sss_i\}_{i=1}^{N_B}$ consisting of $N_B$ category-coherent sketch-image pairs with their semantic representations $\{\sss_i\}$, we denote the hidden state of the $l$-th layer in the multi-modal network of this training batch as $\HH^l$ to be rendered to a graph convolutional layer. As is mentioned in Sec.~\ref{sec312}, for our graph convolutional layers, each training batch is regarded as an $N_B$-vertex graph. Therefore, a convolutional filter $g_\theta$ parameterized by $\theta$ can be applied to $\HH^l$, producing the $(l+1)$-th hidden state $\HH^{(l+1)}=g_\theta\ast\HH^{(l)}$. Suggested by~\cite{gcn}, this can be approached by a layer-wise propagation rule, \ie,
\begin{equation}
	\HH^{(l+1)}=\delta\big(\DD^{-\frac{1}{2}}\AAA\DD^{-\frac{1}{2}}\HH^{(l)}\WW_\theta\big),
\end{equation}
using the first-order approximation of the localized graph filter~\cite{gcn2,hammond2011wavelets}. Again, here $\delta\left(\cdot\right)$ is the activation function and $\WW_\theta$ refers to the linear transformation parameter. $\AAA$ is an $N_B\times N_B$ self-connected in-batch adjacency and $\DD$ can be defined by $\DD=\mathtt{diag\left(\AAA\mathbbm{1}\right)}$. It can be seen in Fig.~\ref{fig2} that the in-batch adjacency $\AAA$ is determined by the semantic representations $\{\sss_i\}$, of which each entry $\AAA^{(j,k)}$ can be computed by $\AAA^{(j,k)}=e^{-\frac{\Vert\sss_j-\sss_k\Vert^2}{t}}$.
In the proposed ZSIH model, two graph convolutional layers are built, with output feature dimensions of $N_B\times 1024$ and $N_B\times M$ for a whole batch. We choose the $\mathtt{ReLU}$ nonlinearity for the first layer and the $\mathtt{sigmoid}$ function for the second one to restrict the output values between 0 and 1.

Intuitively, the graph convolutional layer proposed by~\cite{gcn} can be construed as performing elementary row transformation on a batch of data from fully-connected layer before activation according to the graph Laplacian of $\AAA$. In this way, the semantic relations between different data points are intensified within the network hidden states, benefiting our \zs~hashing model in exploring the semantic knowledge. 
Traditionally, correlating different deep  representations can be tackled by adding a trace-like regularization term in the learning objective. However, this introduces additional hyper parameters to balance the loss terms and the hidden states in the network of different data points are still isolated.
\subsection{Stochastic neurons and decoding network}\label{secsgh}

The encoder-decoder model for ZSIH is introduced in this subsection. Inspired by~\cite{sgh}, a set of latent probability variables $\bb\in(0,1)^M$ are obtained from the second graph convolutional layer output respective to $\{\xx,\yy\}$ corresponding to the hash code for a sketch-image pair $\{\xx,\yy\}$ with the semantic feature $\sss$. The stochastic neurons~\cite{sgh} are imposed to $\bb$ to produce binary codes $\widetilde{\bb}\in\{0,1\}^M$ through a sampling procedure:
\begin{equation}\label{eqb}\vspace{-1ex}
	\widetilde{\bb}^{(m)}=
	\begin{cases}
		1&{\bb^{(m)}\geqslant\epsilon^{(m)},}\\
		0&{\bb^{(m)}<\epsilon^{(m)},}
	\end{cases} \quad\text{for~} m=1~...~M,
\end{equation}
where $\epsilon^{(m)}\sim\mathcal{U}\left([0,1]\right)$ are random variables. As is proved in~\cite{sgh}, this structure can be differentiable, allowing error back-propagation from the decoder to the previous layers. Therefore, the posterior of $\bb$, \ie, $p\left(\bb|\xx,\yy\right)$, is approximated by a Multinoulli distribution:
\vspace{-1.ex}\begin{equation}\vspace{-1ex}
	\begin{split}
		&q(\widetilde{\bb}|\xx,\yy)=\prod_{m=1}^M (\bb^{(m)})^{\widetilde{\bb}^{(m)}}(1-\bb^{(m)})^{1-\widetilde{\bb}^{(m)}}.
	\end{split}
\end{equation}

We follow the idea of generative hashing to build a decoder on the top of the stochastic neurons. During optimization of ZSIH, this decoder is regularized by the semantic representations $\sss$ using the following Gaussian likelihood with the reparametrization trick~\cite{vae}, \ie, 
\vspace{-0.5ex}\begin{equation}\vspace{-0.1ex}
	p\left(\sss|\bb\right)=\mathcal{N}\big(\sss|\mu({{\bb}}), \mathtt{diag}(\sigma^2({{\bb}}))\big),
\end{equation}
where $\mu\left(\cdot\right)$ and $\sigma\left(\cdot\right)$ are implemented by fully-connected layers with identity activations. To this end, the whole network can be trained en-to-end. The learning objective is given in the next subsection.
\subsection{Learning objective and optimization}\label{secobj}
\begin{algorithm}[t]
	\small
	\caption{The Training Procedure of ZSIH}
	\label{alg}
	
	\textbf{Input:}\hspace{0mm} Sketch-image dataset $\mathcal{O}=\{\mathbf{X},\mathbf{Y}\}$, semantic representations $\mathbf{S}$ and max training iteration $T$\\
	\textbf{Output:}\hspace{0mm} Network parameters $\varTheta$\\
	\BlankLine
	\Repeat{convergence or max training iter $T$ is reached}{
		
		Get a random mini-batch $\{\xx_i,\yy_i,\sss_i\}_{i=1}^{N_B}$, assuring $\xx_i,\yy_i$ belong to the same class\\
		
		Build $\AAA$ according to semantic distances\\
		\For{$i=1~...~N_B$}{
			Sample a set of $\epsilon^{(m)}\sim\mathcal{U}\left([0,1]\right)$\\
			Sample a set of $\widetilde{\bb}\sim q(\bb|\xx_i,\yy_i)$}
		$\mathcal{L}\leftarrow$ Eq.~(\ref{eqobj})\\
		$\varTheta\leftarrow\varTheta-\mathbf{\Gamma}\left(\nabla_\varTheta\mathcal{L}\right)$ according to Eq.~(\ref{eqgrad})
	}
	
\end{algorithm}
The learning objective of the whole network for a batch of sketch and image data is defined as follows:
\vspace{-1ex}\begin{equation}\label{eqobj}\vspace{-1ex}
	\begin{split}\vspace{-1ex}
		\mathcal{L}=&\sum_{i=1}^{N_B}\mathbb{E}_{q(\bb|\xx_i,\yy_i)}\big[\log q(\bb|\xx_i,\yy_i)-\log p(\sss_i|\bb)\\
		&+ \frac{1}{2M}\big(\|f(\xx_i) - {\bb}\|^2 + \|g(\yy_i) - {\bb}\|^2\big)\big].
	\end{split}
\end{equation}
Concretely, the expectation term $\mathbb{E}\left[\cdot\right]$ in Eq.~(\ref{eqobj}) simulates the variational-like learning objectives~\cite{vae,sgh} as a generative model. However, we are not exactly lower-bounding any data prior distribution since it is generally not feasible for our ZSIH network. $\mathbb{E}\left[\cdot\right]$ here is an empirically-built loss, simultaneously maximizing the output code entropy via $\mathbb{E}_{q(\bb|\xx,\yy)}[\log q(\bb|\xx,\yy)]$ and preserving the semantic knowledge for the \zs~task by $\mathbb{E}_{q(\bb|\xx,\yy)}[-\log p(\sss|\bb)]$. The single-model encoding functions $f\left(\cdot\right)$ and $g\left(\cdot\right)$ are trained by the stochastic neurons outputs of the multi-modal network using L-2 losses. The sketch-image similarities can be reflected in assigning related sketches and images with the sample code. To this end, $f\left(\cdot\right)$ and $g\left(\cdot\right)$ are able to encode out-of-sample data without additional category information, as the imposed training codes are semantic-knowledge-aware.
The gradient of our learning objective \wrt the network parameter $\Theta$ can be estimated by a Monte Carlo process in sampling $\widetilde{\bb}$ using the small random signal $\epsilon$ according to Eq.~(\ref{eqb}), which can be derived as
\vspace{-1ex}\begin{equation}\label{eqgrad}\vspace{-1ex}
	\begin{split}
		\nabla_\Theta\mathcal{L} \simeq& \sum_{i=1}^{N_B}\mathbb{E}_\epsilon\Big[\nabla_\Theta\Big(\log q(\widetilde{\bb}|\xx_i,\yy_i) - \log p(\sss_i|\widetilde{\bb})\\
		&+\frac{1}{2M}\big(\|f(\xx_i) - \widetilde{\bb}\|^2 +\|g(\yy_i) - \widetilde{\bb}\|^2\big)\Big)\Big].
	\end{split}
\end{equation}
As $\log q(\cdot)$ forms up into an inverse cross-entropy loss and $\log p(\cdot)$ is reparametrized, this estimated gradient can be easily computed. 
Alg.~\ref{alg} illustrates the whole training process of the proposed ZSIH model, where the operator $\mathbf{\Gamma}\left(\cdot\right)$ refers to the Adam optimizer \cite{adam} for adaptive gradient scaling. Different from many existing deep cross-modal and \zs~hashing models \cite{cdq, dsh, zsh, dcmh} which require alternating optimization procedures, ZSIH can be efficiently and conveniently trained end-to-end with SGD.
\subsection{Out-of-sample extension}
When the network of ZSIH is trained, it is able to hash image and sketch data from the \unseen~classes $\mathcal{C}^u$ for matching. The codes can be obtained as follows:
\vspace{-0.1ex}\begin{equation}\label{eqfg}\vspace{-0.1ex}
	\begin{split}
		\mathbf{B}^{im}&=(\operatorname{sign}(\mathit{f}(\mathbf{X}^u-0.5))+1)/2\in\{0,1\}^{N^u\times M}\text{,}\\
		\mathbf{B}^{sk}&=(\operatorname{sign}(\mathit{g}(\mathbf{Y}^u-0.5))+1)/2\in\{0,1\}^{N^u\times M}\text{,}
	\end{split}\vspace{-1ex}
\end{equation}
where $N^u$ is the size of test data. As is shown in Fig.~\ref{fig2}, the encoding networks $f\left(\cdot\right)$ and $g\left(\cdot\right)$ are standing on their own. Semantic representations of test data are not required and there is no need to render data to the multi-modal network. Thus, encoding test data is non-trivial and can be efficient.

\begin{table*}[t]
	\begin{center}
		\caption{\zs~SBIR mAP@all comparison between ZSIH and some cross-modal hashing baselines.}
		\label{tab1}
		\small
		\resizebox{0.9\textwidth}{!}{
			\begin{tabular}{l ccc ccc ccc}
				\hline
				\multirow{2}{*}{\textbf{Method}}&\textbf{Cross}& \textbf{Binary}& \textbf{Zero}&\multicolumn{3}{c}{\textbf{Sketchy (Extended)}}&\multicolumn{3}{c}{\textbf{TU-Berlin (Extended)}}\\\cline{5-10}
				&\textbf{Modal}&\textbf{Code}&\textbf{Shot}& 32 bits & 64 bits & 128 bits & 32 bits & 64 bits & 128 bits\\ \hline\hline
				ZSH~\cite{zsh}&&\checkmark&\checkmark&0.146& 0.165& 0.168& 0.132& 0.139& 0.153\\
				CCA~\cite{cca}&\checkmark&&& 0.092& 0.089& 0.084& 0.083& 0.074& 0.062\\
				CMSSH~\cite{cmssh}&\checkmark& \checkmark& & 0.094& 0.096& 0.111& 0.073& 0.077& 0.080\\
				CMFH~\cite{cmfh}&\checkmark& \checkmark& & 0.115& 0.116& 0.125& 0.114& 0.118& 0.135\\
				SCM-Orth~\cite{scm}&\checkmark& \checkmark& & 0.105& 0.107& 0.093& 0.089& 0.092& 0.095\\
				SCM-Seq~\cite{scm}&\checkmark& \checkmark& & 0.092& 0.100& 0.084& 0.084& 0.087& 0.072\\
				CVH~\cite{cvh}&\checkmark& \checkmark& & 0.076& 0.075& 0.072& 0.065& 0.061& 0.055\\
				SePH-Rand~\cite{seph}&\checkmark& \checkmark& & 0.108& 0.097& 0.094& 0.071& 0.065& 0.070\\
				SePH-KM~\cite{seph}&\checkmark& \checkmark& & 0.069& 0.066& 0.071& 0.067& 0.068& 0.065\\
				DSH~\cite{dsh}&\checkmark& \checkmark& & 0.137& 0.164& 0.165& 0.119& 0.122& 0.146\\
				\hline
				\textbf{ZSIH}&\checkmark&\checkmark&\checkmark&\textbf{0.232}&\textbf{0.254}&\textbf{0.259}&\textbf{0.201}&\textbf{0.220}&\textbf{0.234}\\\hline
			\end{tabular}
		}
	\end{center}
	\vspace{-2ex}
\end{table*}
\begin{figure*}
	\vspace{0ex}\begin{center}
		\includegraphics[width=0.99\textwidth]{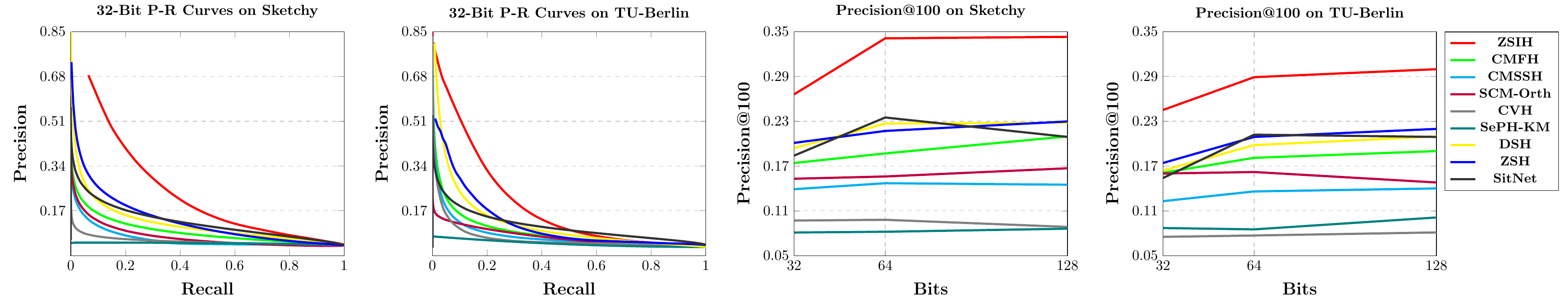}
	\end{center}\vspace{-2ex}
	\caption{Precision-recall curves and precision@100 results of ZSIH and several hashing baselines are shown above. To keep the content concise, only 32-bit precision-recall curves are illustrated here.}
	\label{figpr}
	\vspace{-2ex}
\end{figure*}
\section{Experiments}
\subsection{Implementation details}
The proposed ZSIH model is implemented with the popular deep learning toolbox Tensorflow~\cite{tf}. We utilize the settings of AlexNet~\cite{alexnet} pre-trained on ImageNet~\cite{imagenet} before the last pooling layer to build our image and sketch CNNs. The attention mechanism is inspired by Song~\etal~\cite{sketchatt} without the shortcut connection. The attended 256-D feature is obtained by a weighted pooling operation according to the attention map. All configurations of our network are provided in Fig.~\ref{fig2}. We obtain the semantic representation of each data point using the 300-D word vector~\cite{wv} according to the class name. When the class name is not included in the word vector dictionary, it is replaced by a synonym. For all of our experiments, the hyper-parameter $t$ is set to $t=0.1$ with a training batch size of 250. Our network is able to be trained end-to-end. 
\subsection{Zero-shot experimental settings}
To perform SBIR with binary codes under the novelly-defined \zs~cross-modal setting, the experiments of this work are taken on two large-scale sketch datasets, \ie, Sketchy~\cite{sketchy} and TU-Berlin~\cite{tu}, with extended images obtained from~\cite{dsh}. We follow the SBIR evaluation metrics in \cite{dsh} where sketch queries and image retrieval candidates with the same label are marked as relevant, while our retrieval performances are reported based on nearest neighbour search in the hamming space.

\textbf{Sketchy Dataset~\cite{sketchy} (Extended).} This dataset originally consists of $75,471$ hand-drawn sketches and $12,500$ corresponding images from 125 categories. With the extended $60,502$ real images provided by Liu~\etal~\cite{dsh}, the total size of the whole image set yields $73,002$. We randomly pick 25 classes of sketches and images as the \unseen~test set for SBIR, and data from the rest 100 \seen~classes are used for training. During the test phase, the sketches from the \unseen~classes are taken as retrieval queries, while the retrieval gallery is built using all the images from the \unseen~categories. Note that the test classes are not presenting during training for \zs~retrieval.

\textbf{TU-Berlin Dataset~\cite{tu} (Extended).} The TU-Berlin dataset contains $20,000$ sketches subjected to 250 categories. We also utilize the extended nature images provided in~\cite{dsh, sketchnet} with a total size of $204,489$. 30 classes of images and sketches are randomly selected to form the retrieval gallery and query set respectively. The rest data are used for training. Since the quantities of real images from different classes are extremely imbalanced, we additionally require each test category have at least 400 images when picking the test set.

\subsection{Comparison with existing methods}
\begin{table*}
	\caption{\textit{Zero-shot} sketch-image retrieval performance comparison of ZSIH with existing SBIR and \zs ~learning methods.}
	\label{tab2}
	\small
	\resizebox{0.99\textwidth}{!}{
		\begin{tabular}{cl cccc |cccc}
			\hline
			\multirow{3}{*}{\textbf{Type}}&\multirow{3}{*}{\textbf{Method}}&\multicolumn{4}{c|}{\textbf{Sketchy (Extended)}}&\multicolumn{4}{c}{\textbf{TU-Berlin (Extended)}}\\\cline{3-10}
			&&\textbf{mAP}&\textbf{Precision}&\textbf{Feature}&\textbf{Retrieval}&\textbf{mAP}&\textbf{Precision}&\textbf{Feature}&\textbf{Retrieval}\\
			&&\textbf{@all}&\textbf{@100}&\textbf{Dimension}&\textbf{Time (s)}&\textbf{@all}&\textbf{@100}&\textbf{Dimension}&\textbf{Time (s)}\\\hline\hline
			\multirow{6}{*}{SBIR}& Softmax Baseline & 0.099& 0.176& 4096& $3.9\times10^{-1}$& 0.083& 0.139& 4096&$4.7\times10^{-1}$\\
			& Siamese CNN~\cite{siamese}& 0.143& 0.183& 64& $5.2\times 10^{-3}$& 0.122& 0.153& 64& $6.3\times 10^{-3}$\\
			& SaN~\cite{san}& 0.104& 0.129& 512& $4.4\times 10^{-2}$& 0.096& 0.112& 512& $5.1\times 10^{-2}$\\
			& GN Triplet~\cite{sketchy}& 0.211& 0.310& 1024& $8.9\times10^{-2}$& 0.189& 0.241& 1024& $1.4\times 10^{-1}$\\
			& 3D Shape~\cite{3dshape}& 0.062& 0.070& 64& $5.6\times 10^{-3}$&0.057& 0.063& 64& $7.0\times 10^{-3}$\\
			& DSH (64 bits)~\cite{dsh}& 0.164& 0.227& 64 (binary)& $6.3\times 10^{-5}$& 0.122& 0.198& 64 (binary)& $7.5\times 10^{-5}$\\\hline
			\multirow{7}{*}{\textit{Zero-Shot}}& CMT~\cite{cmt}& 0.084& 0.096& 300& $3.1\times 10^{-2}$& 0.065& 0.082& 300& $3.7\times 10^{-2}$\\
			& DeViSE~\cite{devise}& 0.071& 0.078& 300& $3.2\times 10^{-2}$& 0.067& 0.075& 300& $3.7\times 10^{-2}$\\
			& SSE~\cite{sse}& 0.108& 0.154& 100& $1.1\times 10^{-2}$& 0.096& 0.133& 220& $1.3\times 10^{-2}$\\
			& JLSE~\cite{jlse}& 0.126& 0.178& 100& $1.1\times 10^{-2}$& 0.107& 0.165& 220& $1.3\times 10^{-2}$\\
			& SAE~\cite{sae}& 0.210& 0.302& 300& $3.1\times 10^{-2}$& 0.161& 0.210& 300& $3.7\times 10^{-2}$\\
			& ZSH (64 bits)~\cite{zsh}& 0.165& 0.217& 64 (binary)& $6.3\times 10^{-5}$& 0.139& 0.174& 64 (binary)& $7.5\times 10^{-5}$\\\hline
			\textbf{Proposed}& \textbf{ZSIH (64 bits)}& \textbf{0.254}& \textbf{0.340}& 64 (binary)& $6.5\times 10^{-5}$& \textbf{0.220}& \textbf{0.291}& 64 (binary)& $7.9\times 10^{-5}$\\\hline
		\end{tabular}
	}
	\vspace{-1ex}
\end{table*}

As cross-modal hashing for SBIR under the \zs~setting has never been proposed before to the best of our knowledge, the quantity of potential related existing baselines is limited. Our task can be regarded as a combination of  conventional cross-modal hashing, SBIR and \zs~learning. Therefore, we adopt existing methods according to these themes for retrieval performance evaluation. We use the \seen-\unseen~splits identical to ours for training and testing the selected baselines. The deep-learning-based baselines are retrained end-to-end using the \zs~setting mentioned above. For the non-deep baselines, we extract the respective AlexNet~\cite{alexnet} \texttt{fc\_7} features pre-trained on the \seen~sketches and images as model training inputs for a fair comparison with our deep model.

\begin{table}
	\caption{Ablation study. 64-bit mAP@all results of several baselines are shown below.}
	\label{tab3}
	\small
	\resizebox{0.99\linewidth}{!}{
		\begin{tabular}{l cc}
			\hline
			\textbf{Description}& \textbf{Sketchy}& \textbf{TU}\\\hline\hline
			Kron. layer $\rightarrow$ concatenation& 0.228& 0.207\\
			Kron. layer $\rightarrow$ MFB~\cite{mfb}& 0.236& 0.211\\
			Stochastic neuron $\rightarrow$ bit regularization& 0.187& 0.158\\
			Decoder $\rightarrow$ classifier& 0.162& 0.133\\
			Without GCNs& 0.233& 0.171\\
			GCNs $\rightarrow$ word vector fusion& 0.219& 0.176\\
			$t=1$ for GCNs& 0.062& 0.055\\
			$t=10^{-6}$ for GCNs& 0.241& 0.202\\\hline
			\textbf{ZSIH (full model)}& \textbf{0.254}& \textbf{0.220}\\\hline
		\end{tabular}
	}
	\vspace{-3ex}
\end{table}

\textbf{Cross-Modal Hashing Baselines.} Several state-of-the-art cross-modal hashing works are introduced including CMSSH~\cite{cmssh}, CMFH~\cite{cmfh}, SCM~\cite{scm}, CVH~\cite{cvh}, SePH~\cite{seph} and DSH~\cite{dsh}, where DSH~\cite{dsh} can also be subjected to an SBIR model and thus is closely related to our work. In addition, CCA~\cite{cca} is considered as a conventional cross-modal baseline, though it learns real-valued joint representations.

\textbf{\textit{Zero-Shot}~Baselines.} Existing \zs~learning works are not originally designed for cross-modal search. We select a set of state-of-the-art \zs~learning algorithms as benchmarks, including CMT~\cite{cmt}, DeViSE~\cite{devise}, SSE~\cite{sse}, JLSE~\cite{jlse}, SAE~\cite{sae} and the \zs~hashing model, \ie, ZSH~\cite{zsh}. For CMT~\cite{cmt}, DeViSE~\cite{devise} and SAE~\cite{cmt}, two sets of 300-D embedding functions are trained for sketches as images with the word vectors~\cite{wv} as the semantic information for nearest neighbour retrieval, and the classifiers used in these works are ignored. SSE~\cite{sse} and JLSE~\cite{jlse} are based on \seen-\unseen~class mapping, so the output embedding sizes are set to 100 and 220 for Sketchy~\cite{sketchy} and TU-Berlin~\cite{tu} dataset respectively. We train two modal-specific encoders of ZSH~\cite{zsh} simultaneously for our task.

\textbf{Sketch-Image Mapping Baselines.} Siamese CNN~\cite{siamese}, SaN~\cite{san}, GN Triplet~\cite{sketchy}, 3D Shape~\cite{3dshape} and DSH~\cite{dsh} are involved as SBIR baselines. We follow the instructions of the original papers to build and train the networks under our \zs~setting. A softmax baseline is additionally introduced, which is based on computing the 4096-D AlexNet~\cite{alexnet} feature distances pre-trained on the \seen~classes for nearest neighbour search.

\begin{figure}[t]
	\begin{center}
		\includegraphics[width=0.99\linewidth]{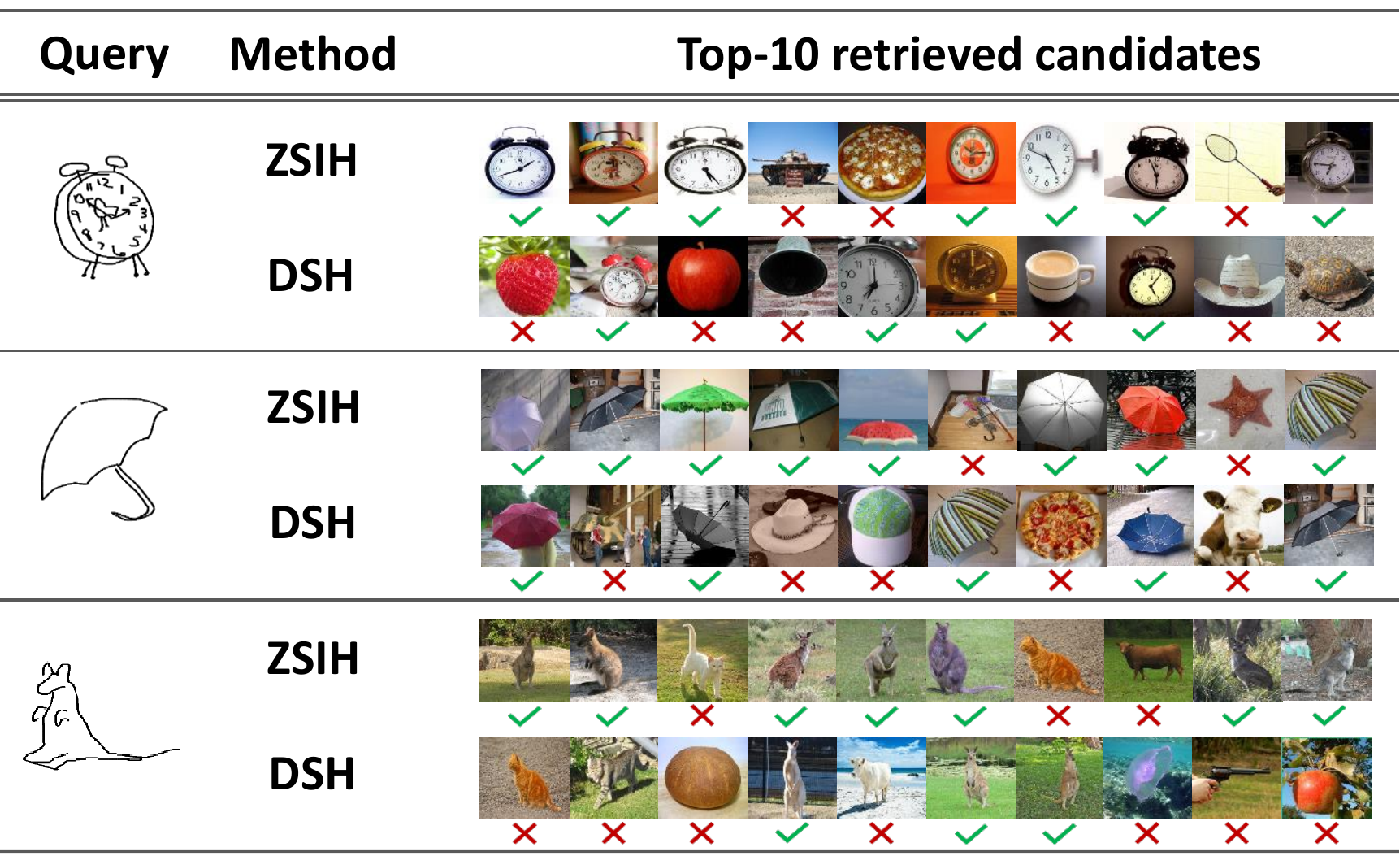}
	\end{center}\vspace{-1ex}
	\caption{Some top-10 \zs~SBIR results of ZSIH and DSH~\cite{dsh} are shown here according to the hamming distances, where the green ticks indicate correct retrieval candidates and red crosses indicate the wrong ones.}
	\label{figsee}
	\vspace{-1ex}
\end{figure}

\begin{figure}[t]\vspace{-0ex}
	
	\begin{center}
		\includegraphics[width=0.99\linewidth]{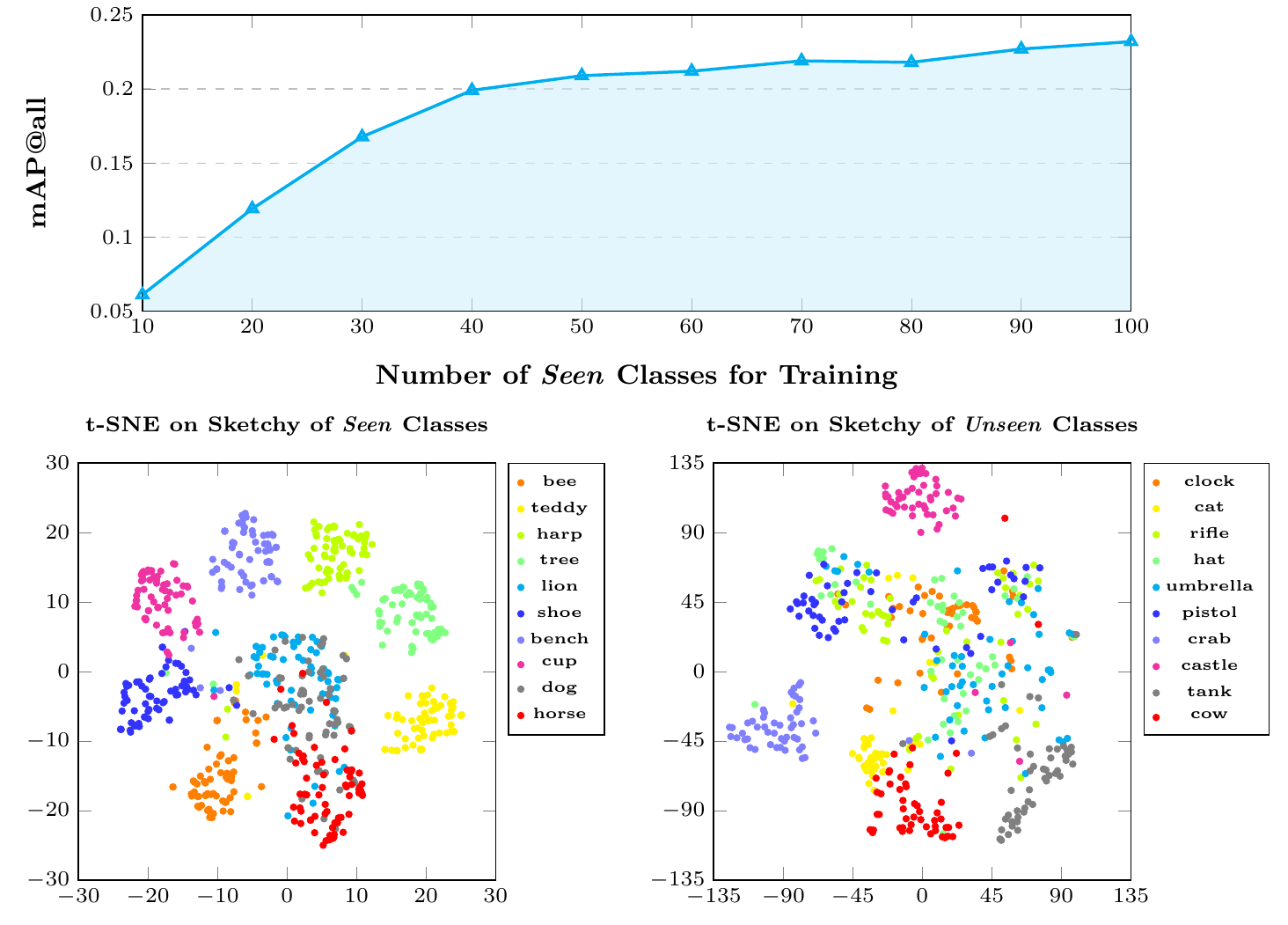}
	\end{center}
	\vspace{-2ex}
	\caption{First row: 32-bit ZSIH retrieval performance on Sketchy according to different numbers of \seen~classes used during training. Second row: 32-bit t-SNE~\cite{tsne} scattering results on the Sketchy dataset of the \seen~and \unseen~classes.}\vspace{-1ex}
	\label{figtsne}
\end{figure}

\textbf{Results and Analysis.} The \zs~cross-modal retrieval mean-average precisions (mAP@all) of ZSIH and several hashing baselines are given in Tab.~\ref{tab1}, while the corresponding precision-recall (P-R) curves and precision@100 scores are illustrated in Fig.~\ref{figpr}. The performance margins between ZSIH and the selected baselines are significant, suggesting the existing cross-modal hashing methods fail to handle our \zs~task. 
ZSH~\cite{zsh} turns out to be the only well-known \zs~hashing model and it attains relatively better results than other baselines. However, it is originally designed for single-modal data retrieval. 
DSH~\cite{dsh} leads the SBIR performance under the conventional cross-modal hashing setting, but we observe a dramatic performance drop when extending it to the \unseen~categories.  
Some retrieval results are provided in Fig.~\ref{figsee}. Fig.~\ref{figtsne} shows the 32-bit t-SNE~\cite{tsne} results of ZSIH on the training set and test set, where a clearly scattered map on the \unseen~classes can be observed. We also illustrate the retrieval performance \wrt the number of \seen~classes in Fig.~\ref{figtsne}. It can be seen that ZSIH is able to produce acceptable retrieval performance as long as an adequate number of \seen~classes is provided to explore the semantic space.

The comparisons with SBIR and \zs~baselines are shown in Tab.~\ref{tab2}, where an akin performance margin to the one of Tab.~\ref{tab1} can be observed. To some extent, the SBIR baselines based on positive-negative samples, \eg, Siamese CNN~\cite{siamese} and GN Triplet~\cite{sketchy}, have the ability to generalize the learned representations to \unseen~classes. SAE~\cite{sae} produces closest performance to ZSIH among the \zs~learning baselines. Similar to ZSH~\cite{zsh}, these \zs~baselines suffer from the problem of mitigating the modality heterogeneity. Furthermore, most of the methods in Tab.~\ref{tab2} learn real-valued representations, which leads to poor retrieval efficiency when performing nearest neighbour search in the high-dimensional continuous space.

\subsection{Ablation study}

Some ablation study results are reported in this subsection to justify the plausibility of our proposed model.

\textbf{Baselines.} The baselines in this subsection are built by modifying some parts of the original ZSIH model. To demonstrate the effectiveness of the Kronecker layer for data fusion, we introduce two baselines by replacing the Kronecker layer~\cite{kron} with the conventional feature concatenation and the multi-modal factorized bilinear pooling (MFB) layer~\cite{mfb}. Regularizing the output bits with quantization error and bit decorrelation loss identical to ~\cite{dh} is also considered as a baseline in replacing the stochastic neurons~\cite{sgh}. The impact of the semantic-aware encoding-decoding design is evaluated by substituting a classifier for the semantic decoder. We introduce another baseline by replacing the graph convolutional layers~\cite{gcn} with conventional fully connected layers. Fusing the word embedding to the multi-modal network is also tested in replacement of graph convolution. Several different hyper-parameter settings of $t$ are also reported.

\textbf{Results and Analysis.} The ablation study results are demonstrated in Tab.~\ref{tab3}. We only report the 64-bit mAP on the two datasets for comparison in order to ensure the paper content to be concise. It can be seen that the reported baselines typically underperform the proposed model. Both feature concatenation and MFB~\cite{mfb} produce reasonable retrieval performances, but the figures are still clearly lower than our original design. We speculate this is because the Kronecker layer considerably expands the hidden state dimensionality and therefore, the network is able to store more information for cross-modal hashing. When testing the baseline of bit regularization similar to \cite{dh}, we experience an unstable training procedure easily leading to overfitting. The quantization error and bit decorrelation loss introduce additional hyper-parameters to the model, making the training procedure hard. Replacing the semantic decoder with a classifier results in a dramatic performance fall as the classifier basically provides no semantic information and fails to generalize knowledge from the \seen~classes to the \unseen~ones. Graph convolutional layer~\cite{gcn} also plays an important role in our model. The mAP drops by about $4\%$ when removing it. Graph convolution enhances hidden representations and knowledge within the neural network by correlating the data points that are semantically close, benefiting our \zs~task. As to the hyper-parameters, a large value of $t$, \eg, $t=1$, generally leads to a tightly-related graph adjacency, making data points from different categories hard to be recognized. On the contrary, an extreme small value $t$, \eg, $t=10^{-6}$, suggests a sparsely-connected graph with binary-like edges, where only data points from the same category are linked. This is also suboptimal in exploring the semantic relation for \zs~tasks.

\section{Conclusion}
In this paper, a novel but realistic task of efficient large-scale \textit{zero}-\textit{shot} SBIR hashing was studied and successfully tackled by the proposed \zs~sketch-image hashing (ZSIH) model. We designed an end-to-end three-network deep architecture to learn shared binary representations and encode sketch/image data. Modality heterogeneity between sketches and images was mitigated by a Kronecker layer with attended features. Semantic knowledge was introduced in assistance of visual information by graph convolutions and a generative hashing scheme. Experiments suggested the proposed ZSIH model significantly outperforms existing methods in our \zs~SBIR hashing task.

{\small
	\bibliographystyle{ieee}
	\bibliography{ymbib}
}
\end{document}